\documentclass[sigconf]{acmart}

\AtBeginDocument{%
  \providecommand\BibTeX{{%
    \normalfont B\kern-0.5em{\scshape i\kern-0.25em b}\kern-0.8em\TeX}}}

\setcopyright{acmcopyright}
\copyrightyear{2021}
\acmYear{2021}
\setcopyright{acmcopyright}
\acmConference[ICMR '21]{Proceedings of the 2021 International Conference on Multimedia Retrieval}{August 21--24, 2021}{Taipei, Taiwan}
\acmBooktitle{Proceedings of the 2021 International Conference on Multimedia Retrieval (ICMR '21), August 21--24, 2021, Taipei, Taiwan}
\acmPrice{15.00}
\acmDOI{10.1145/3460426.3463635}
\acmISBN{978-1-4503-8463-6/21/08}

\begin{document}

\title{Relation-aware Hierarchical Attention Framework for Video Question Answering}

\author[*]{Fangtao Li}
\email{	lift@bupt.edu.cn}
\author{Ting Bai}

\email{baiting@bupt.edu.cn}
\affiliation{%
	\institution{Beijing University of Posts and Telecommucations}
	%\streetaddress{P.O. Box 1212}
	\city{Beijing}
	\country{China}
}
\author{Chenyu Cao}
\email{www.caochenyu@bupt.edu.cn}
\author{Zihe Liu}

\email{ziheliu@bupt.edu.cn}

\affiliation{%
	\institution{Beijing University of Posts and Telecommucations}
	\streetaddress{P.O. Box 1212}
	\city{Beijing}
	\country{China}
}

\author{Chenghao Yan}
\email{yanch@bupt.edu.cn}
\author{Bin Wu}
\authornotemark[2]
\email{wubin@bupt.edu.cn}
\affiliation{%
	\institution{Beijing University of Posts and Telecommucations}
	\streetaddress{P.O. Box 1212}
	\city{Beijing}
	\country{China}
}

\renewcommand{\shortauthors}{Anonymous Author}
\fancyhead{}

\renewcommand{\shortauthors}{Li, et al.}

\begin{abstract}
	
	Video Question Answering (VideoQA) is a challenging video understanding task since it requires a deep understanding of both question and video. Previous studies mainly focus on extracting sophisticated visual and language embeddings, fusing them by delicate hand-crafted networks.
	However, the relevance of different frames, objects, and modalities to the question are varied along with the time, which is ignored in most of existing methods.
	Lacking understanding of the the dynamic relationships and interactions among objects brings a great challenge to VideoQA task.
	To address this problem, we propose a novel Relation-aware Hierarchical Attention (RHA) framework to learn both the static and dynamic relations of the objects in videos. In particular, videos and questions are embedded by pre-trained models firstly to obtain the visual and textual features. Then a graph-based relation encoder is utilized to extract the static relationship between visual objects.
	To capture the dynamic changes of multimodal objects in different video frames, we consider the temporal, spatial, and semantic relations, and fuse the multimodal features by hierarchical attention mechanism to predict the answer. We conduct extensive experiments on a large scale VideoQA dataset, and the experimental results demonstrate that our RHA outperforms the state-of-the-art methods. 
	%To capture the dynamic changes of multimodal objects in different video frames, we consider three dynamic relations: i.e., the temporal relations, spatial relations, and semantic relations, and fuse the multimodal features by hierarchical attention mechanisms to predict the answer. We conduct extensive experiments on a large scale VideoQA dataset, and the experimental results demonstrate that our RHA outperforms the state-of-the-art methods. 

\end{abstract}



\maketitle
\begin{figure}[tb]
	\includegraphics[width=\linewidth]{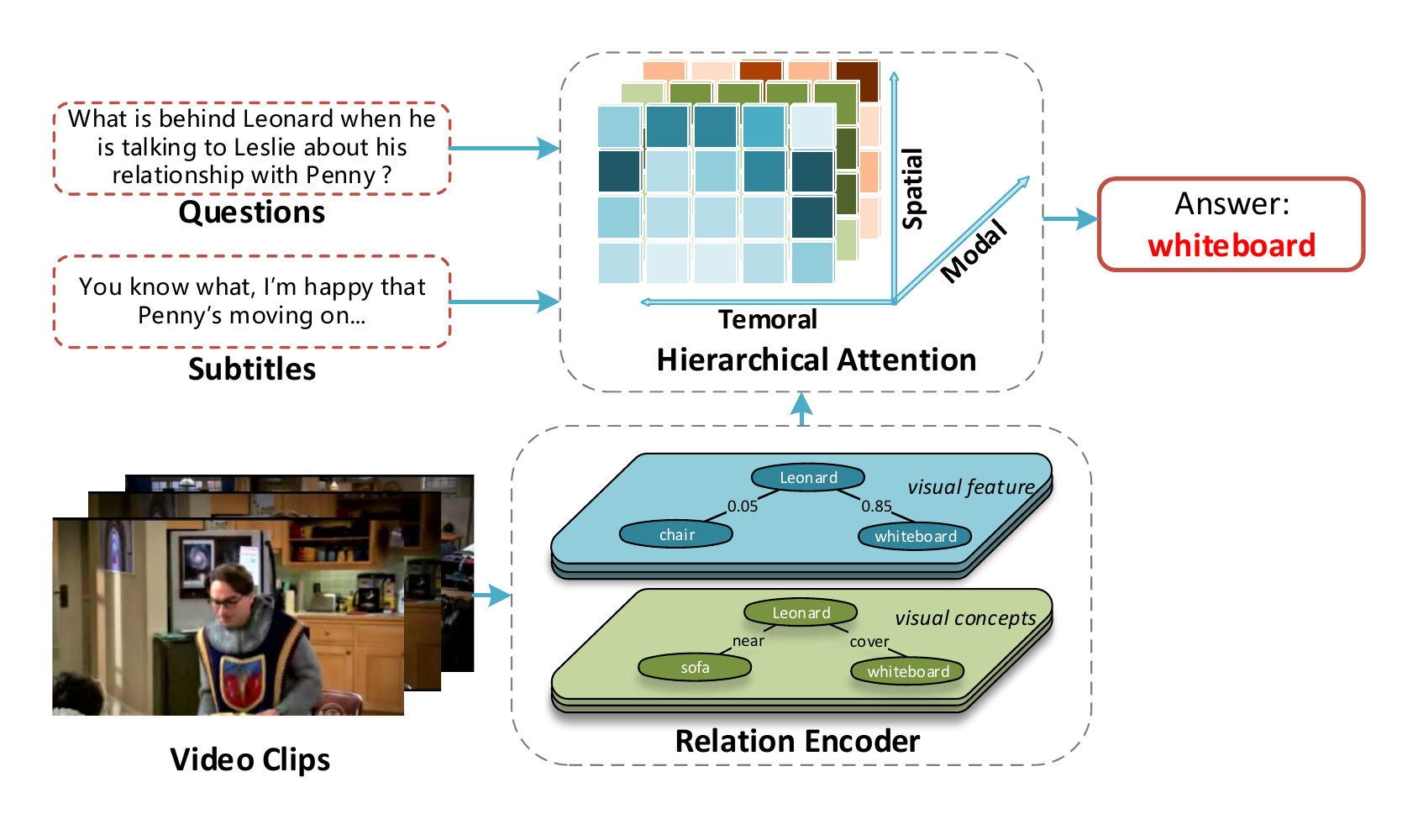}
	\caption{The illustration of Relation-aware Hierarchical Attention framework.} \label{Fig1}
\end{figure}

\section{Introduction}
With the rapid development of deep learning in Computer Vision (CV) and Natural Language Processing (NLP).
Video Question Answering (VideoQA), as a interdisciplinary area between language and vision, has been an active research topic in recent studies in video understanding.
Given a video clip and a video-related question, VideoQA is expected to answer the question based on video grounding, reasoning, and translating. As a practical and comprehensive task, VideoQA can benefit other video understanding tasks such as video retrieval and storytelling.

The general VideoQA architecture consists of a detector, an embedding module, and a predictor. 
Usually, objects in the video are firstly detected and then fed into a embedding module. The embedding modules mainly utilize the Convolutional Neural Network (CNN) or pre-trained R-CNN~\cite{girshick2015fast} to obtain the visual features of objects in video, and sequential models such as Long Short-Term Memory (LSTM) to encode the texts in the questions. After aligning the question and video by attention mechanism or bilinear pooling~\cite{kim2018bilinear, ben2017mutan}, the multimodal features of both video and question are jointly learned to obtain the answer to the question by the predictor.

However, most of these methods lack the ability of question localization in videos, which is vital for getting the proper representations to answer the question. 
In temporal, only some key frames are closely related to the question. In spatial, only crucial areas and objects are useful to answering. 
Hence, both the static alignment in a single frame and the dynamic alignment among all frames in the video between the question and objects are significant. 
Besides, a video consists of multimodal features, i.e., visualizations, conceptions, and subtitles, and the intra-relations of different modalities should also been taken into consideration. 

To address the above problems,  we propose a novel Relation-aware Hierarchical Attention (RHA) framework to learn both the static and dynamic relations of the visual objects in videos.
For the static relations of visual objects in a single frame, we construct a graph to build the spatial and semantic relationships among objects. Then we measure the modality importance by taking their relevance to the question into consideration. For example, in Figure~\ref{Fig1}, given a question as "\textit{What is behind Leonard when he is talking to Leslie about his relationship with Penny?}", previous methods~\cite{lei2020tvqa+, lei2018tvqa} incorporate all information into modeling, including the relationship between "Leslie" and "Penny". While such relationship is usefulness and what we focused is the object behind "Leonard", which can been learned from the visual information of video. The importance of different modalities should be considered at the static feature fusion stage.
As for learning the dynamic changes of visual objects in different video frames, we characterize the temporal, spatial, and semantic relations. %Considering different dynamic relations may have different effects to the answer of the question, 
We adopt attention mechanism to learn from each dynamic relations.
Our proposed RHA model learn the static relation among objects, as well as the dynamic relations in time, space and modality dimensions. On the temporal dimension, the hierarchical attention locates essential time stamps. On the spatial dimension, it identifies important areas, and in the modality dimension, the importance of different modality features can also be learned. 

The architecture of our framework is shown in Figure~\ref{fig2}. A pre-trained BERT~\cite{devlin2018bert} is applied to extract the features of questions, candidate answers, and subtitles. For each frame in the video, a pre-trained Faster R-CNN~\cite{ren2015faster} is applied for object detection and embedding. %Each bounding box is also be annotated. 
All the objects are then fed into a customized graph attention networks (GATs)~\cite{velivckovic2017graph} with shared weights as a relation encoder to learn the relationships between objects. The updated embeddings will be injected into a multimodal attention module, along with subtitle and question embeddings. Finally, we use a Multi-Layer-Perceptron (MLP) as an answer prediction module to generate the correct answer index.
We evaluate our framework on TVQA+~\cite{lei2020tvqa+} dataset and conduct experiments on temporal grounding, showing the the ability of our model in pointing out relevant temporal span. 

To summarize, our main contributions are:

\begin{itemize}
	\item We propose a novel RHA framework which adopts a question-guided hierarchical attention module to capture both static and dynamic relations of multimodal objects.
	
	\item We introduce a graph-based relation encoder to model the static relationships between visual objects in videos, and three dynamic relations: temporal relation, spatial relations, and semantic relations are considered to predict the answer. 
	\item We verify the proposed framework RHA on a large scale TVQA+ dataset. Experimental results show that our model outperforms other state-of-the-art methods.
\end{itemize}

\section{Related Works}
\subsection{Video Question Answering}
%%%%%%%%%%%%%%%%%%%%%%%%%%%%%%%%%%%%%
As a typical multimodal task, VideoQA requires thorough visual and textual understanding. In recent years, some more restricted sub-tasks have also been proposed to enhance the interpretability, such as Knowledge-based VideoQA~\cite{garcia2020knowit} and Spatio-temporal grounding VideoQA~\cite{lei2020tvqa+}. Nevertheless, the VideoQA framework generally consists of a video encoder, a question encoder, an embedding alignment module, and a predictor. With the domination of deep learning, early methods~\cite{gao2018motion, zeng2017leveraging} use CNN to extract features at frame level. However, most of the CNN-based extractors cannot seize temporal information in the video. ~\cite{yu2017end, jiang2020reasoning} apply LSTM as a substitute to model temporal context. As for question encoder, Glove~\cite{pennington2014glove} and LSTM are generally applied. %In recent years, with the great success of BERT~\cite{devlin2018bert}, some recent works~\cite{lei2020tvqa+} also migrate pre-trained transformers-based models to the question embedding domain. 
As the kernel of the whole VideoQA framework, embedding alignment module could be quite sophisticated. Early works~\cite{zeng2017leveraging} rely on hand-craft CNN architecture to further embed video and question. Inspired by ~\cite{vaswani2017attention}, ~\cite{huang2020location} utilizes attention-based methods to focus on relevant video clips. Attention mechanism also enhances the interpretability of these models, since it works in a simple but intuitive way. Another line of research focuses on graph-based learning. With the popularity of Graph Neural Networks (GNNs)~\cite{kipf2016semi}, some works focus on modeling video from the perspective of topology and embed video by GCNs. ~\cite{jiang2020reasoning} constructs a heterogeneous graph which regards both words in the question and frames in the video clip as nodes and aligns them by GCNs~\cite{kipf2016semi}.

The most related work is STAGE model~\cite{lei2020tvqa+}, which proposes a VideoQA framework with spatial and temporal grounding. Compared to STAGE, our RHA contains some critical distinctions: (1) STAGE designs an elaborate convolutional kernel for encoding, which is highly customized and cannot capture any relation attributes. In comparison, RHA utilizes a graph-based encoder to learn both objects embedding and their relations; (2) STAGE fuses multimodal features by a heuristic method, while we propose an interpretable hierarchical attention module to fuse multimodal features adaptively.

\subsection{Relation Understanding}
%%%%%%%%%%%%%%%%%%%%%%%%%%%%%%%%%%%%%
%Relation understanding is initially introduced as a task of Natural Language Processing (NLP), but it is also developing flourishingly in the Computer Vision field. 
Relation understanding contains many sub-tasks, in which relation extraction and relation reasoning are two of the most important task. Relation extraction aims to detect relationships between given objects. Prior works recognize relations between objects by co-occurrence~\cite{shih2016look} and position~\cite{yao2018exploring}. This kind of methods is generally based on statistics and can only identify spatial relations (such as \textit{behind, below}, and \textit{cover}). Another line of work is semantic relation extraction, which generally requires a deeper understanding of video. ~\cite{hu2018relation} proposes a neural network to extract semantic relations on a single image. ~\cite{dai2019two} designs a novel Two-Stage Model to extract the social relationships between characters. Undeniably, as a stepping-stone of video understanding, most relation extraction methods are designed as a task-specific module. ~\cite{cadene2019murel} builds a trainable cell named MuRel to model pair-wise object relationships, while ~\cite{hu2018relation} discovers implicit relation by adopting an attention-based object relation detector.

In contrast, relation reasoning aims to represent objects based on their relations. ~\cite{le2020hierarchical} proposes a Relation Network as a general solution of relation reasoning in an unsupervised manner. ~\cite{sun2020learning} designs a novel Interaction Canonical Correlation Network for cross-modal relation reasoning.  With the explosive development of GNNs, recent research suggests that objects and their relationship can be represented by nodes and edges in the graph. ~\cite{santoro2017simple} builds a fully-connected graph for a given image, discovering interactions with a self-attention mechanism.

Our work is also inspired by ~\cite{li2019relation}, which builds a relation-aware graph network to discover explicit and implicit relations in the image. We make some targeted improvements for VideoQA task. The first difference is that we explore objects relations in a video rather than a single image. Second, we use both visual features and visual concepts for relation encoding, which enhances the interpretability and robustness of our framework.

\begin{figure*}
	\includegraphics[width=\textwidth]{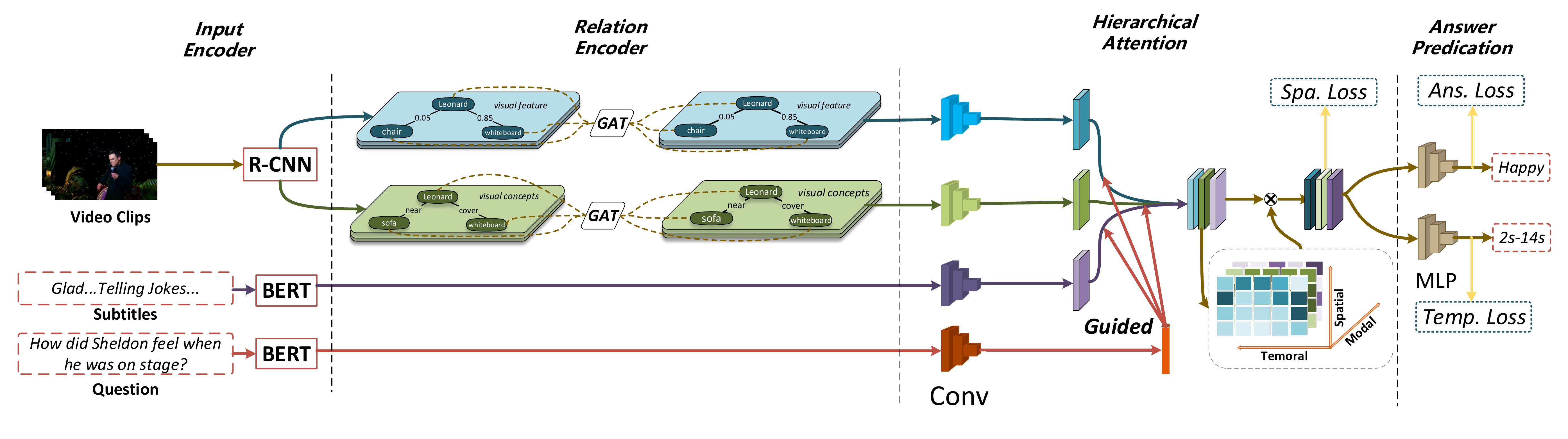}
	\caption{An overview of Relation-aware Hierarchical Attention framework } \label{fig2}
\end{figure*}
\subsection{Multimodal Fusion}

As one of the original topics in multimodal learning, multimodal fusion aims at gaining joint representation of two or more modalities. Some studies~\cite{wang2020multi, lei2018tvqa, lei2020tvqa+} applied vector operations between single-modal features, including vector concatenation, element-wise multiplication, and element-wise addition. Such researches are referred to as early fusion since they fuse multimodal information before the decision. In contrast, late fusion uses unimodal decision results and merges the results by a fusion mechanism such as averaging or voting~\cite{baltruvsaitis2018multimodal}. With the development of deep learning, some works~\cite{ramirez2011modeling,glodek2011multiple} have proposed to use trainable model to enhance the performance of multimodal fusion. As for unsupervised learning, ~\cite{padua2019multimodal} presents a multimodal autoencoder to fuse features adaptively without any supervision. Network Architecture Search (NAS) is also applied in multimodal fusion~\cite{perez2019mfas}. More recently, bilinear pooling is a practical pathway of fusion~\cite{kim2018bilinear}. ~\cite{zadeh2017tensor} calculates the outer-product of video, acoustic and textual features to gain the multimodal joint representation. ~\cite{liu2018efficient} proposes a low-rank method to build tensor networks to reduce computational complexity caused by tensor outer-product. With the domination of bilinear pooling, attention mechanism ~\cite{zadeh2018memory} is also regarded as an effective method to enhance the interaction between modalities and avoid redundancy. ~\cite{liu2018iqiyi} proposes a Multimodal Attention (MMA) module, which reweights modalities through the Gram matrix, and ~\cite{li2021frame} optimizes MMA by design a deeper attention convolutional layer. 

In terms of multimodal fusion, ~\cite{jiang2020reasoning} and ~\cite{lei2020tvqa+} are the most similar works. However, ~\cite{lei2020tvqa+} simply assumes that each modality has the same weight, ignoring the difference between questions. Co-attention proposed by ~\cite{jiang2020reasoning} lacks the ability to analyze fine-grained object-level importance. To the best of our knowledge, we are the first to measure the importance of different time, objects, and modalities at the same time in this task.
\section{Methods}
\subsection{The General Framework of RHA}
Our work mainly focuses on multiple-choice VideoQA task, which needs to choose the right answer in a set of candidate answers. Given (1) a question $q$; (2) 5 candidate answers $\{a_{k} | k = 1, ... , 5 \}$; (3) a video clip that consists of keyframes $\{ F_{t} | t = 1, ... , T\}$ (4) the subtitles of corresponding video $\{ P_{t} | t = 1, ... , T\}$, our goal is to predict the index of right answer $\hat{a}$:
\begin{equation}
	\hat{a} = \mathop{\arg\max}_{a \in a_{k}}p(a|q,P,F).
\end{equation}

The RHA framework consists of four modules. As shown in Figure \ref{fig2}, all visual and text inputs are first embedded by the input encoder. Then a relation encoder is utilized to discover relationships between visual objects. All these representations are projected into the same dimension after relation modeling. To learn the relevance between video and question, we apply the hierarchical attention module to reweight and fuse representations. Finally, we predict the right answer and its relevance video clip by an answer predictor.

\subsection{Input Encoder}
For a given video containing keyframes $\{F_t\}$, we use Faster R-CNN to extract the features of each detected object, followed by PCA to downsize the dimension of object proposals into a low dimension representation $\textbf{O} = \{\textbf{o}_t^i \in \mathbb{R}^{d_o}, i \le N_o, t \le T\}$, where ${\textbf{o}_t^i}$ refers to the \textit{i}-th objects of \textit{t}-th frame. The bounding-boxes $\textbf{B} = \{ \textbf{b}_t^i \in \mathbb{R}^{4}, i \le N_o, t \le T\}$ are represented by the coordinate of the top-left and bottom-right point of the bounding-boxes. The label of objects is embedded to a vector $ \textbf{L} = \{ \textbf{l}_t^i \in \mathbb{R}^{d_l}, i \le N_o, t \le T\}$ by Glove~\cite{pennington2014glove}, where $d_l$ is set to 300. For subtitles $\{ P_t\}$, we use a pre-trained BERT to extract embeddings $ \textbf{S} = \{ \textbf{S}_t \in \mathbb{R}^{L_s \times d_s}, t \le T\}$. The candidate answers are first concatenated with the question to compose a qa-hypothesis $\{\textbf{Q}_{k}\}$. The same pre-trained BERT is used to extract embeddings which are denoted as $ \textbf{H} = \{ \textbf{h}_k \in \mathbb{R}^{L_q \times d_q}, k \le 5\}$, where $d_s$ and $d_q$ denote the dimension of word embedding, while $L_s$ and $L_q$ refer to the length of subtitles and hypothesis, respectively.
\subsection{Relation Encoder}
Given a video clip, we discover explicit (spatial) and implicit (semantic) relationships between different objects in the video. Understanding these relationships is the key to understanding the video underlying information. To capture these relationship attributes in the video, we build an encoder based on GATs~\cite{velivckovic2017graph} to capture the relationship between objects, processing the spatial relationship and semantic relationship, respectively. 

%\subsection*{Spatial Relation}
\subsubsection{Spatial Graph Construction}
Spatial relation in the video refers to the position relationship between objects. The spatial relation may be varied with the variance of the position of objects and movement of camera. The frame-level spatial relation is denoted as a graph $\textbf{G}^t_{spa}=(\textbf{V}^t_{spa},\textbf{E}^t_{spa})$, and the video-level spatial relation can be represented as the concatenation of $\textbf{G}^t$. We denote $\textbf{v}^{t,i}_{spa}$ as the embedding of \textit{i}-th node in frame \textit{t}, and $\textbf{e}_{spa}^{t,i,j}$ refers to the spatial relation between object \textit{i} and \textit{j}. Inspired by ~\cite{yao2018exploring}, the spatial relations between objects are classified into 11 categories based on bounding-boxes $\textbf{B}_t$. Each category refers to a kind of spatial relations such as \textit{cover, in,} and \textit{near}. Note that we use the label of objects $\{\textbf{l}_t \in \mathbb{R}^{N_o \times d_l} \}$ as node embeddings, rather than the visual features of objects. We argue that alignment from words in question to objects in video is the key for understanding spatial relation, while the visual features such as shape, color is irrelevant. And it is more difficult to align visual features to text features than to align text features, which affects the accuracy of answering questions.

\subsubsection{Spatial Graph Update}
For each keyframe in the video, we get a spatial graph $\textbf{G}^t_{spa}$ using the methods above. Then we utilize a customized GAT~\cite{velivckovic2017graph} to update the node embedding. We adopt multi-head attention to generalize the learning progress of GAT. All output features of heads are concatenated. In order to encode spatial relation into GAT, the bias is set to be independent for different spatial relations, and the projection matrix represents the direction of relation (\textit{from objects} or \textit{to objects}):
\begin{equation}
	\mathbf{l}_i = ||_{m = 1}^{M}\sigma ( \sum_{j \in \mathcal{N}_i} \alpha_{ij}^{m} \cdot \mathbf{W}_{spa}^{m} \mathbf{l}_{j} ),
\end{equation}
\begin{equation}
	\alpha_{ij} = \frac{exp((\mathbf{U}_{spa}\mathbf{l}_{i})^\top \cdot \mathbf{V}_{spa}^{dir(i,j)}\mathbf{l}_{j} + \mathbf{b}_{spa}^{lab(i,j)})}{\sum_{j \in \mathcal{N}_i} exp((\mathbf{U}_{spa}\mathbf{l}_{i})^\top \cdot \mathbf{V}_{spa}^{dir(i,j)}\mathbf{l}_{j} + \mathbf{b}_{spa}^{lab(i,j)})},
\end{equation}
where $\alpha \in \mathbb{R}^{N_o \times N_o}$ is the attention weight, $\textbf{W}_{spa} \in \mathbb{R}^{M \times d_h \times d_l} $, $\textbf{U}_{spa} \in \mathbb{R}^{d_h \times d_l}$, $\textbf{V}_{spa}^{ dir } \in \mathbb{R}^{ d_h \times d_l}$ is the projection matrix, ${\mathbf{b}_{spa}^{lab}} \in \mathbb{R}^{d_h} $ is the bias, and $d_h$ is the dimension of hidden layer. $M$ refers to the number of heads of graph attention, which is set to 15 in our implementation. Residual connection is also involved to avoid over smoothing in GAT. Updated frame-level features $\textbf{v}_{spa}^{t}$ can be represented as the concatenation of node embeddings.

For different frames in the video, we share the parameters of GAT. One advantage of weight sharing is that it can reduce the number of parameters significantly. On the other hand, this relation encoder can be more robust to the temporal changing of spatial relationship and the changing of the number of objects.

\subsubsection{Semantic Graph Construction}
Semantic relation in the video refers to the relationship that can not be inferred only by postion and visual information. Similar to spatial relation, semantic relation between objects may be changed along with the progress of plots. Given visual features $\{\textbf{o}_t \in  \mathbb{R}^{N_o \times d_l} \}$, the frame-level semantic relation between objects $\textit{i}$ and $\textit{j}$ in frame $\textit{t}$ is defined as below.
\begin{equation}
	\mathbf{e}_{sem}^{t, i,j} = \frac{exp(\textbf{W}_s [\textbf{o}_t^i; \textbf{o}_t^j])} { \sum_{j \in \mathcal{N}_i} exp(\textbf{W}_s [\textbf{o}_t^i; \textbf{o}_t^j])},
\end{equation}
where $\textbf{W}_s \in \mathbb{R}^{2d_o \times 1}$ is trainable parameters. Treating each object $\{\textbf{o}_t^i\}$ as a node, we build the semantic graph $\textbf{G}^t_{sem}=(\textbf{V}^t_{sem},\textbf{E}^t_{sem})$ where $\textbf{e}_{sem}^{t,i,j}$ refers to the semantic relation between object \textit{i} and \textit{j}. Note that in this stage, we use the region features $\{O_t^i\}$ as the embedding of nodes since the semantic relation between objects is hard to infer only by their categories. More visual information is involved as supplementary of relation understanding.
\subsubsection{Semantic Graph Update}
Following the previous works~\cite{li2019relation}, we utilize graph attention mechanism to update the node embedding after building semantic graph $\textbf{G}_t^{sem}$ for each keyframe in the video. Multi-head attention is also applied in GAT. All output features of attention heads are concatenated to obtain the updated node embeddings:
\begin{equation}
	\mathbf{o}_i = ||_{m = 1}^{M}\sigma ( \sum_{j \in \mathcal{N}_i} \beta_{ij}^{m} \cdot \mathbf{W}_{sem}^{m} \mathbf{o}_{j} ),
\end{equation}
\begin{equation}
	\beta_{ij} = \frac{exp((\mathbf{U}_{sem}\mathbf{o}_{i})^\top \cdot \mathbf{V}_{sem}\mathbf{o}_{j})}{\sum_{j \in \mathcal{N}_i} exp((\mathbf{U}_{sem}\mathbf{o}_{i})^\top \cdot \mathbf{V}_{sem}\mathbf{o}_{j})},
\end{equation}
where $\beta_{ij} \in \mathbb{R}^{N_o \times N_o}$ is the attention weight, $\textbf{W}_{sem} \in \mathbb{R}^{d_h \times d_o} $, $\textbf{U}_{sem} \in \mathbb{R}^{d_h \times d_o}$, $\textbf{V}_{sem} \in \mathbb{R}^{d_h \times d_o}$ is the projection matrix, and $m$ is the number of heads of graph attention, which is set to 15 in our implementation. For semantic graph, we do not encode relation direction, since the semantic graph is fully-connected and symmetric. Residual connection is also involved to avoid over smoothing in GAT. For different frames in the video, we also share the parameters of GAT for semantic graph update. Similar to spatial relation, each frame can be represented as the concatenation of node embeddings $\textbf{v}^{sem} = ||_{t=1}^{T} \textbf{o}_t^{sem}$.

\subsection{Hierarchical Attention Module}
As stated before, a video consists of visual feature, visual concepts, and subtitles. In order to measure the relevance to the question, the visual and textual features will be downsized into the same dimension first. Following the previous works~\cite{lei2018tvqa, lei2020tvqa+}, we adopt a fully-connected layer with residual connection to build downsize encoding block for object features $\{\textbf{o}_t\}$, layer normalization~\cite{ba2016layer} is also involved:
\begin{equation}
	\mathbf{o}_{t}^i = Layernorm(ReLU(\mathbf{W}_d\mathbf{o}_{t}^i) + \mathbf{o}_{t}^i).
\end{equation}
Similar procedure is applied on visual concepts $\{ \textbf{l}_t\}$, subtitles $ \{\textbf{s}_t\}$, and qa-hypothesis $\{\textbf{h}_k\}$. Then the question is grounded in temporal, spatial, and modal, generating an updated video features with the question encoded. Note that in our model, we distinguish the visual features and visual concepts, since although they all represent visual information, the description methods are different. 
\subsubsection{Spatial and Temporal Attention}
At the first stage, we locate the qa-hypothesis  spatially and temporally. Given the encoded hypothesis $\{\mathbf{h}_k\}$ and encoded visual features $\{\mathbf{o}_t\}$, the attention scores of visual features $M_{k,t} \in \mathbb{R}^{L_q \times N_o}$ and visual representation $\mathbf{o}_{t,att} \in \mathbb{R}^{L_q \times d_h}$ is computed as:
\begin{equation}
	M_{k,t} = softmax(\mathbf{h}_{k} \cdot \mathbf{o}_{t}^{\top}),
\end{equation}
\begin{equation}
	\mathbf{o}_{t, att} = M_{k,t} \mathbf{o}_{t}.
\end{equation}
The question-guided attention scores actually represent the relevance between the qa-hypothesis and the objects in different frames so that the predictor can focus on important visual objects. The same process is performed on visual concepts and subtitles to compute representation $\mathbf{l}_{t, att} \in \mathbb{R}^{L_q \times d_h}$ and $\mathbf{s}_{t, att} \in \mathbb{R}^{L_q \times d_h}$. 

\subsubsection{Multimodal Attention}
In the ideal case, all modalities are as same important for answering the question. However, the questions are often changeable and comprehensive, resulting in the redundancy of video. Different questions may only focus on the information of a certain modal in the video. In this case, we need to focus on some modalities to answer the question better. Motivated by this, we design a multimodal attention mechanism at the second stage of the hierarchical attention. First, we concatenate $\mathbf{o}_{t,att}$, $\mathbf{l}_{t,att}$, and $\mathbf{s}_{t,att}$ as multimodal features $X_t \in \mathbb{R}^{ L_q \times d_h \times 3}$. The feature $X_t$ is transformed into low dimension space by trainable parameters $\textbf{W}_F \in \mathbb{R}^{\widehat{d}_h \times d_h}$. Then we adopt the Gram matrix of $X_t$ to capture modal correlation by multiplying $X_t$ with its transpose. The weight of each modal is yielded through the convolution layer and a softmax activation: 
\begin{equation}
	Y_i = \sum_{j=1}^{3}X_{j}\cdot \frac{exp{((\mathbf{W}_{F}X)^{\top} (\mathbf{W}_{F}X))_{j,i}}}{\sum_{i}{exp{((\mathbf{W}_{F}X)^{\top} (\mathbf{W}_{F}X))_{j,i}}}}.
\end{equation}

Note that at the second stage, we apply a self-attention mechanism for multimodal fusion rather than question guided attention, since the question has been encoded into the multimodal features in the first stage so that the second stage mainly focuses on the importance of the modalities themselves.

\subsection{Answer Predictor}
The last module is an answer and localization predictor. The RHA is required to predict the answer based on multimodal features $Y \in \mathbb{R}^{T \times L_q \times d_h}$. Minimum time spans related to the question are also predicted based on the joint representation. We first apply a convolutional layer with max-pooling layer to obtain the output $A \in \mathbb{R}^{T \times d_h}$. For temporal prediction, $A \in \mathbb{R}^{T \times d_h}$ is sent into two linear layers with softmax to produce start probabilities $\textbf{p}_{k}^{1} \in \mathbb{R}^T$ and end probabilities $\textbf{p}_{k}^{2} \in \mathbb{R}^T$ for each frame $k$. For answer prediction, an additional linear layer is first utilized to further encode video-text representation $A_k$. Then a global representation $G_k^g \in \mathbb{R}^{d_h}$ is generated by max-pooling across all the time steps. Taking temporal prediction into consideration, we generate temporal proposals using dynamic programming. For each proposal, we generate a local representation $G_k^{l}\in \mathbb{R}^{d_h}$ by max-pooling $A_k$, concatenating with $G_k^g$ to obtain $G \in \mathbb{R}^{2d_h \times 5}$ . Finally, concatenated features $G$ is sent to a softmax function to generated the answer scores $\textbf{P}^{ans} \in \mathbb{R}^5$. Note that we mainly implemented this module based on STAGE~\cite{lei2020tvqa+}, and we only make some essential changes to fit the input of visual features, visual concepts, and subtitles for a fair comparison.

\subsection{Training Loss}
As we need to answer the question with temporal grounding, the RHA framework is trained with supervision from ground truth (GT) bounding boxes, GT time proposal, and GT answer. For spatial supervision, we define a box as positive for spatial prediction if it has an IoU larger than 0.5 with the GT box. The attention weights of positive objects should be higher than negative ones. So LSE~\cite{li2017improving} loss function is applied since it is easier to optimize~\cite{bartlett2008classification}:
\begin{equation}
	loss_{Spa} = \frac{1}{N}\sum_{i=1}^{N} \sum_{r_p \in \Omega_p, r_n \in \Omega_n}^{N} log(1+exp(M_{i,t,r_n} - M_{i,t,r_p})),
\end{equation}
where $M_{i,t,r}$ is the $r$-th element of the matching scores $M_{i,t}$. For temporal supervision, cross-entropy loss is applied to measure the probabilities of start and end time:
\begin{equation}
	loss_{Temp} = \frac{1}{2N} \sum_{i=1}^{N}({y_i^{st}}log\textbf{P}^1 + {y_i^{ed}}log\textbf{P}^2),
\end{equation}
where $y^{st}$ and $y^{ed}$ is ground truth start and end indices. Similarly, given answer probabilities $\textbf{P}^{ans}$, we also apply a cross-entropy loss as answer prediction loss:
\begin{equation}
	Loss_{Ans} = \frac{1}{N} \sum_{i=1}^{N} {y_i^{ans}}log\textbf{P}^{ans},
\end{equation}
where $y^{ans}$ is the index of ground truth answer. Finally, all three losses are summed up as:
\begin{equation}
	Loss = \omega_{Ans} Loss_{Ans} + \omega_{Spa} loss_{Spa} + \omega_{Temp} loss_{Temp},
\end{equation}
where $\omega_{Ans}$, $\omega_{Spa}$ and $\omega_{Temp}$ are the weights of different loss. In our case, $\omega_{Ans}$ is set to 1, while $\omega_{Spa}$ and $\omega_{Temp}$ are both 0.5.
\section{Experiment}

\subsection{Dataset}
TVQA+~\cite{lei2020tvqa+} is a large scale multiple-choice VideoQA dataset with spatio-temporal grounding. All data is collected from "The Big Bang Theory". The TVQA+ dataset is the augmented version of TVQA dataset~\cite{lei2018tvqa}, with 21.8K 60-90 seconds long video clips and 29.4K multiple-choice questions grounded in both the temporal and the spatial domains. Each question is followed by 5 candidate answers, in which only one of them is the right answer. For spatio-temporal grounding, there are 310.8K bounding boxes linked with referred objects and people, spanning across 2.5K categories. All questions are composed of a question part("\textit{Where/Why/What}") and a localization part("\textit{when/before/after}"). 
%The original TVQA+ dataset is split into train set, valid set, and test set randomly according to the ratio of 8:1:1. %Annotations of the test set are not released to avoid data leakage.

\subsection{Implementation Details}
In our implementation, Layer Normalization~\cite{ba2016layer} and Dropout is applied between every two full-connected layers. 
%We apply ReLU as non-linear activation functions for most of the layers except the last one, of which softmax is applied to generate scores for all candidate answers. 
As for hyper-parameters, dimension of object features $d_o$ is set to 300, while textual feature dimension $d_s$ and $d_q$ are 768. The dimension of hidden layers $d_h$ is set to 128, while $\widehat{d}_h$ in hierarchical attention is 32. The dropout rate is set to 0.1. At the training stage, batch size is set to 16 to balance the training speed and memory cost, while at the inference stage, we set batch size to 1. Adam optimizer is applied for training, the initial learning rate is set to 0.001 and will be decayed by 0.1 for every 10 epochs.

\subsection{Evaluation Metrics}

In this work, we use three evaluation metrics to measure performance. First, we use classification accuracy to measure QA performance. We also consider temporal localization performance, which is evaluated by temporal mean Intersection-over-Union (Temp. mIoU). Finally, we evaluate QA accuracy and temporal localization jointly by Answer-Span joint Accuracy (ASA)~\cite{lei2020tvqa+}. For this metric, we regard a prediction as positive only if the predicted temporal localized span has an IoU $\ge$ 0.5 with the ground-truth span and the answer is correctly predicted at the same time.

\subsection{Experimental Results}
We mainly compare with state-of-the-art methods on TVQA+~\cite{lei2020tvqa+}. ST-VQA~\cite{jang2017tgif} is designed for question answering on short videos or GIFs. Two-stream model~\cite{lei2018tvqa} is a method to predict the answer based on videos and subtitles, respectively. The two-stream model is retrained based on official code and TVQA+ data since the original two-stream uses Glove rather than BERT~\cite{devlin2018bert} to embed subtitles, which may result in worse performance. STAGE~\cite{lei2020tvqa+} also encodes frame-wise regional visual representations and neural language representations, which also implements temporal localization and spatial grounding. STAGE-sub means only the subtitle branch is activated for question answering, while STAGE-vid means only video features are taken as input. STAGE is also retrained with the official code. All models are trained on the train set and tested on the test set. The experimental results are shown in Table~\ref{tab1}. We select STAGE as our main baseline method. Experimental results show that our RHA model outperforms other methods by a significant margin(74.34 \textit{vs.} 72.14). 

We also report the performance of our model on different question types. As shown in Table~\ref{tab2}, we classify the questions according to the first word and select the most frequent five classes. 
%ASA and Temp. mIoU do not fluctuate much on all types of questions. This suggests that RHA could locate relevant video clips precisely. 
Surprisingly, our RHA performs quite well for questions started with "Why", which are generally regarded as hard cases in VideoQA. This phenomenon indicates that RHA captures implicit information for reasoning and answering. However, for questions started with "Who" or "How", RHA still leaves a lot to be desired, which may be the aftermath of the absence of acoustic analysis.
\begin{table}[tb]
	\caption{Comparison with state-of-the-art methods on TVQA+ test set.}\label{tab1}
	\begin{center}
		\begin{tabular}{c|c|c|c}
			\hline
			Model&Acc &  Temp. mIoU & ASA \\
			\hline
			
			ST-VQA~\cite{jang2017tgif}&48.28 &- &-\\
			two-stream~\cite{lei2018tvqa}&68.13 &- &-\\
			STAGE-video~\cite{lei2020tvqa+}&52.75 & 10.90 &2.76\\
			STAGE-sub~\cite{lei2020tvqa+}&67.99 &30.16 &20.13\\
			STAGE~\cite{lei2020tvqa+}&72.14 &30.68 &20.99\\
			FAMF(Ours)&\textbf{74.34} &\textbf{31.53} &\textbf{21.77}\\
			
			\hline
		\end{tabular}
	\end{center}
\end{table}
\begin{table}[tb]
	\caption{Evaluation by question type on TVQA+ valid set}\label{tab2}
	
	\begin{center}
		\begin{tabular}{c|c|c|c}
			\hline
			Model&Acc &  Temp. mIoU & ASA \\
			\hline
			
			What& 72.23 &31.68 &20.81\\
			Why&81.60 &31.81 &21.75\\
			Where&73.63 & 31.38 &23.97\\
			Who&69.57 &26.96 &15.21\\
			How&69.23 &30.68 &20.99\\	
			\hline
		\end{tabular}
	\end{center}
	
\end{table}
\begin{table}[tb]
	\caption{Experimental results of using different modalities on valid set.}\label{tab3}
	
	\begin{center}
		\begin{tabular}{l|c|c|c}
			\hline
			model &  Acc  & Temp. mIoU & ASA  \\
			\hline
			baseline & 70.16& 30.05 & 19.52  \\
			RHA-re &  71.72&  30.43 &  19.98  \\
			RHA-vc & 71.89 &  30.76 &  20.11 \\
			RHA-vs & 72.08 &  30.88 &  20.15  \\
			RHA-ha & 72.45 &  30.93 &  20.31 \\
			RHA-full & \textbf{72.58}&  \textbf{31.30} & \textbf{20.64} \\
			\hline
		\end{tabular}
	\end{center}
	
\end{table}
\begin{figure*}
	\includegraphics[width=\textwidth]{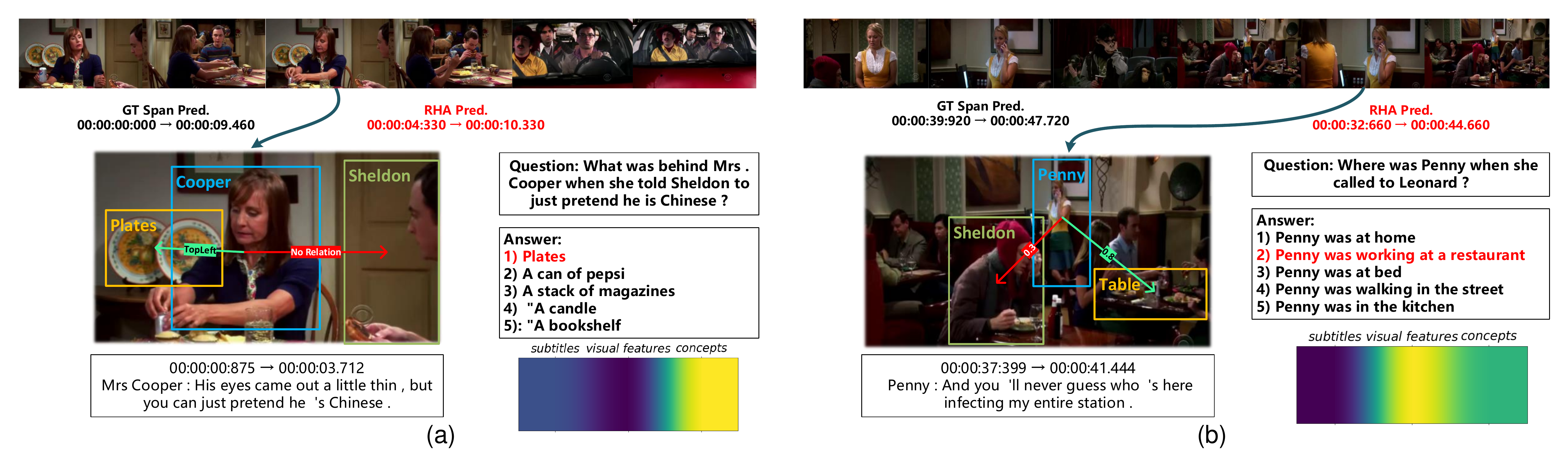}
	\caption{Positive examples to illustrate the process of the RHA framework. Darker color refers to lower modality weight.} \label{figure3}
\end{figure*}
\subsection{Ablation Study}
To measure the effectiveness of our proposed RHA framework, we drop the relation encoder and hierarchical attention, respectively. We report the ablation results on the TVQA+ valid set, which are shown in Table~\ref{tab3}. As stated before, we select STAGE as our main baseline method. RHA-re means we only applied relation encoder and no attention fusion is activated. RHA-vc and RHA-vs mean we use visual concepts and visual features for relation encoder, respectively. RHA-ha means only the hierarchical attention module is applied, and the relation encoder is dropped. RHA-full is the full version of RHA framework.
\subsubsection{The effect of Relation Encoder}
First, we analyze the effect of the relation encoder. From \textit{lines} 1 and 2 in Table~\ref{tab3}, we can find the relation encoder brings an accuracy improvement of about 1\% compared to the baseline model. The results show that introducing relation modeling, which can discover the potential relationships between objects, helps better video understanding. For semantic relation, commonsense (such as traffic rules, physics rules, etc.) is less considered in the previous VideoQA models, and this knowledge can be described in terms of the internal connections between objects. For example, given a question like "\textit{Why Sheldon stop the car}", we need to build a connection between traffic lights and vehicles, answering the question based on the fact that the traffic light is red. These kinds of questions are exactly what the previous models difficult to learn by convolutional layer, but can be learned through our implicit relation module.

On the other hand, some questions may refer to the position of objects. Similarly, taking "\textit{What is near Sheldon}" as a toy example, previous models may use large amounts of training data for pattern recognition, answering the question based on objects that often appear with Sheldon. However, these methods do not really understand the concept of "near", but only choose the candidate answer based on the principle that "\textit{what you see is what you answer}". Our spatial relation module encodes the position of objects by their bounding boxes so that the RHA can explicitly modeling spatial relationships and answer these kinds of questions from an interpretable perspective.

\subsubsection{The usefulness of Visual Concepts and Visual Features}
One difference between the RHA framework and previous models is that we use visual concepts rather than visual features to model spatial relations. Compared to visual features, visual concepts are more abstract and concise as a text-based category. The 2\textit{nd} and 3\textit{rd} lines in Table~\ref{tab3} show that using visual concepts for the spatial relation module can improve performance. The results validate our assumption that the abstract features that are easy to align with the question are more suitable than the specific features that are not in the same space as questions.
In the ideal case that every object mentioned in question has its annotations, we can focus on understanding spatial relationship, ignoring the bias caused by object alignment.

As shown in the 2\textit{nd} and 4\textit{th} lines in Table~\ref{tab3}, accuracy will decrease when the visual concepts are applied in the semantic relation encoder. We argue that using only categories in implicit relation reasoning is too hard. The absence of visual semantic information leads to wrong or unreasonable reasoning. For example, it is hard to judge why a vehicle is stopped if we only know there is a traffic light but do not know its direction and color. Compared to using too abstract embedding that may lead to misunderstanding, it is a better solution to use more detailed visual features in the reasoning stage and then use a convolutional layer or a fusion module to perform the alignment.

\subsubsection{The effect of Hierarchical Attention}
At last, we discuss the influence of the Hierarchical Attention (HA) module. \textit{Lines} 1 and 5 in Table~\ref{tab3} shows that HA brings an improvement of about 2\% in accuracy. In the first stage of HA, the question is used to measure the weight of frames and subtitles temporally. In general, only part of the time periods is related to the questions. By magnifying the weight of important time periods, the RHA framework can discover information better. Although in these periods, only a few objects and words in subtitles are strongly related (\textit{e.g.} some words are cited in the question, or the question mentions specific objects). Thus the scores of each words and objects are calculated. The question is actually encoded into the representation of the video in this stage so that the video representation is adapted to the specific questions.

In the second stage of HA, we reweight different modalities. One disadvantage of the question-guided spatio-temporal attention mentioned in the first stage is that all modalities actually have the same contribution for question answering. However, it is common that the question is unrelated to some modalities (\textit{e.g.} the problem may be purely visually related and not related to subtitles). Previous research lacks a discussion on the importance of modality. According to the multimodal attention mechanism, our RHA framework learns the weight of different modalities adaptively, reaching a more precise localization of questions. Another scenario like "What does Sheldon say..." is also a typical case. Although it refers to a person, what did the person say is more important. For these questions, the key point is searching in candidate answers that have a similar representation with subtitles. 

%We believe the HA module mimics the process of question answering as human beings. Given a certain question, people always first roughly locate the relevant time period in the video based on the time positioning word of the question. Second, people will find the mentioned objects and words according to the question and focus on them. Last, if the topic of the question is about visual information such as color and shape, then humans will focus on the visual objects.
\subsection{Case Study}

To illustrate the RHA framework performance better, we select two right-answered examples randomly to better illustrate the process of RHA framework. As shown in Figure \ref{figure3}, we visualize the main objects of each frame and their relations. The weight of modalities is also visualized by color. Figure ~\ref{figure3}(a) shows the influence of spatial modeling. Given the question as "What was behind Mrs. Cooper when she told Sheldon to just pretend he is Chinese?", ground truth time annotation is 0\textit{s}-9.46\textit{s}, and the true answer is "Plates". At the 3.71 seconds of the video, the subtitles mentioned, "\textit{Mrs. Cooper: His eyes came out a little thin, but you can just pretend he's Chinese}". 
%Through our RHA framework, we locate the start point of this scene at 4.33\textit{s}, which only differs from the relevant subtitles for about 0.5 seconds. 
The difference between ground truth and our result (0\textit{s} \textit{vs.} 4.33\textit{s}) may because before 4.33\textit{s}, although Cooper and Sheldon had occurred, Cooper did not say anything, so the RHA regards these frames as irrelevant. After 10\textit{s}, plates never appeared again, and the conversation between Sheldon and Cooper shifted to other topics. RHA labeled 10.33\textit{s} as the end time, which
% is less than one second away from the ground truth and 
is still in the acceptable error range. As for spatial relation modeling, two main objects related to "\textit{Mrs. Cooper}" are "\textit{Sheldon}" and "\textit{plates}". Among them, Sheldon was sitting on the right side in \textit{frame} 2 and 4, and the plates were first overlapped with Cooper at \textit{frame} 1, then at \textit{frame} 3, the spatial relation between them is recognized as "\textit{close to the left}". Based on the spatial adjective "\textit{behind}" in the question, RHA infers "\textit{Plates}" as the right answer, indicating that RHA can understand spatial relations in video.

Figure~\ref{figure3}(b) is another example. The given question is "\textit{Where was Penny when she called to Leonard?}", time annotation is from 39.92\textit{s} to 47.72\textit{s}, while the ground truth answer is "\textit{Penny was working at a restaurant.}", main object annotations include Penny, Leonard, and some tableware. RHA proposed 32.66\textit{s}-42.66\textit{s} as temporal localization. We find that at 32\textit{s}, Sheldon was sitting in the restaurant, and at 42\textit{s}, Penny called to Leonard.
%During this period, the restaurant and the cinema appeared alternately, while the ground truth start point (39.92\textit{s}) is the first time Penny appeared. 
This case shows that RHA is inclined to use the moment of scene transition as the key point. 
%Plus, subtitles did not indicate the location obviously, so Penny's location needs to be inferred from the objects in the scene. 
RHA caught some objects that only appeared in the restaurant and connected them to Penny with a higher implicit relation score to answer the question. We also find our RHA gives subtitles a lower weight, which proves that RHA could focus on specific modalities by Hierarchical Attention.

\begin{figure}[tb]
	\includegraphics[width=\columnwidth]{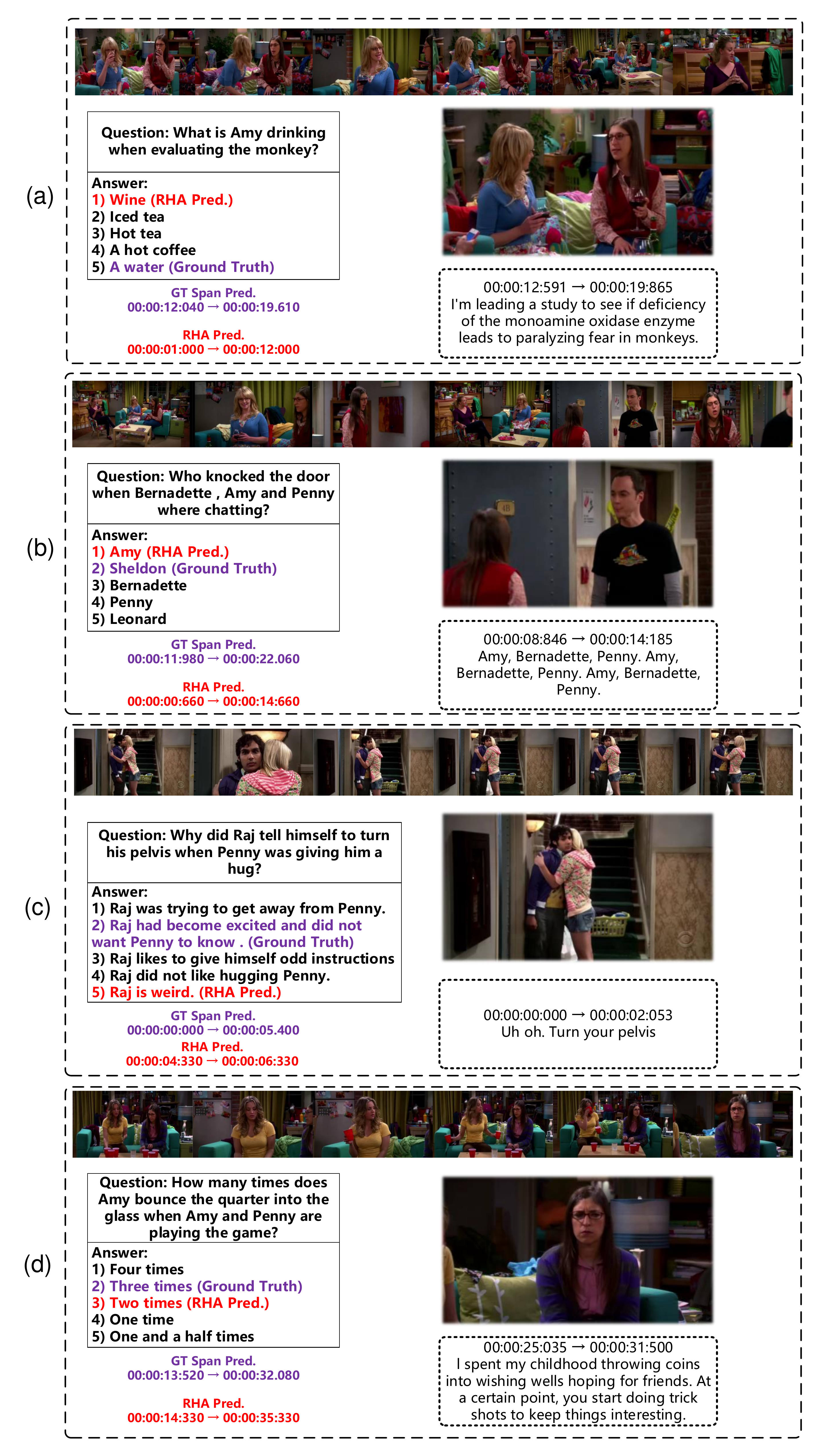}
	\caption{Some negative examples of TVQA+ valid set. RHA predictions are colored in red, while the ground truth predictions are colored in purple.} \label{figure4}
\end{figure}

Plus, we also visualize some negative examples predicted by our RHA framework, as shown in Figure~\ref{figure4}. Most of the negative cases can be classified into four categories: (1)Temporal ambiguity; (1)Audio understanding; (2)Causal reasoning; (3)Counting. Figure~\ref{figure4}(a) shows a typical case of temporal ambiguity. For given question "\textit{What is Amy drinking when evaluating the monkey?}", we need to understand the meaning of "\textit{evaluating}", which is obscurely hidden in Amy's dialogue. In Figure~\ref{figure4}(b), although the question "\textit{Who knocked the door when Bernadette, Amy, and Penny were chatting?}" is not quite difficult, the moment of knocking requires listening to the sound. In case that RHA only takes keyframes and subtitles as input, it can not understand any acoustic information, which finally leads to time dislocating (0-14\textit{s vs.} 11-22\textit{s}). In Figure~\ref{figure4}(c), the question "\textit{Why did Raj tell himself to turn his pelvis when Penny was giving him a hug?}" can be answered by the fact that Raj likes Penny and he is glad to have physical contact with her. However, commonsense and causal reasoning need external knowledge as a supplement, which is not involved in our framework. For these questions, knowledge-based methods provide a feasible idea. In Figure~\ref{figure4}(d), we show a negative example caused by counting error. Given a question "\textit{How many times does Amy bounce the quarter into the glass when Amy and Penny are playing the game?}", RHA locates the event precisely (14.33\textit{s}-35.66\textit{s} \textit{vs.} 13.52\textit{s}-32.08\textit{s}). However, RHA fails to count the times of bouncing. For temporal counting, there is not a well-performed method yet, since it needs a deep understanding of action and number.
\section{Conclusion}
In this paper, we propose a novel RHA framework for VideoQA task. As a challenging video understanding task, VideoQA needs a comprehensive understanding of both visual and textual information. To address the video redundant phenomenon, we design a novel hierarchical attention module. The hierarchical attention module firstly measures the temporal and spatial importance based on their relevance to the question. Then scores of different modalities are calculated at the second stage to fuse multimodal features efficiently. As an important underlying semantic information, relations between objects reflect interaction and connections. To involve relation understanding into VideoQA,  We build a graph-based relation encoder to capture such relation information and embed it into objects by weight-shared GATs. Experimental results on TVQA+ dataset validate the performance of RHA.
% We expect our RHA framework can bring improvement to video-textual understanding. In the future, we will involve external knowledge and acoustic analysis into RHA to enhance its performance.

\begin{acks}
This work is supported by the National Natural Science Foundation of China under Grant No.61972047, the National Key Research and Development Program of China (2018YFC0831500), the Fundamental Research Funds for the Central Universities (500420824), the NSFC-General Technology Basic Research Joint Funds under Grant U1936220 and the Fundamental Research Funds for the Central Universities (2019XD-D01).
\end{acks}

\newpage
\bibliographystyle{ACM-Reference-Format}
\bibliography{acmart}

\end{document}